\documentclass[twoside,a4paper,11pt]{article}
\usepackage{graphicx}
\usepackage{caption}
\usepackage{longtable}
\usepackage{array}
\usepackage{geometry}
\usepackage{bm}
\usepackage{wrapfig}

\usepackage{times}
\usepackage{elcvia}
\usepackage[english]{babel}
\usepackage{hyperref}

\input{psfig.sty}

\title{A Labeled Array Distance Metric for Measuring Image Segmentation Quality
\footnote[0]{Correspondence to: berijani@msu.edu \vspace{0.2cm}}
}

\author{Maryam Berijanian$^1$, Katrina Gensterblum$^1$, Doruk Alp Mutlu$^1$, \\Katelyn Reagan$^{2,3}$, Andrew Hart$^1$, Dirk Colbry$^1$\\ \\
\centerline{\small \em $^1$ SEE-Insight Research Lab, Michigan State University, East Lansing, MI, USA} \\
\centerline{\small \em $^2$ University of Wisconsin–Madison, Madison, WI, USA} \\ 
\centerline{\small \em $^3$ Smith College, Northampton, MA, USA} \\\\
       }

\begin{document}

\maketitle

\pagestyle{myheadings}

\markboth{\centerline{\small \it M. Berijanian et al.}}
         {\centerline{\small \it M. Berijanian et al.}}
\hrulefill

\begin{abstract}

This work introduces two new distance metrics for comparing labeled arrays, which are common outputs of image segmentation algorithms. Each pixel in an image is assigned a label, with binary segmentation providing only two labels (`foreground' and `background'). These can be represented by a simple binary matrix and compared using pixel differences. However, many segmentation algorithms output multiple regions in a labeled array. We propose two distance metrics, named LAD and MADLAD, that calculate the distance between two labeled images. By doing so, the accuracy of different image segmentation algorithms can be evaluated by measuring their outputs against a `ground truth' labeling. Both proposed metrics, operating with a complexity of $O(N)$ for images with $N$ pixels, are designed to quickly identify similar labeled arrays, even when different labeling methods are used. Comparisons are made between images labeled manually and those labeled by segmentation algorithms. This evaluation is crucial when searching through a space of segmentation algorithms and their hyperparameters via a genetic algorithm to identify the optimal solution for automated segmentation, which is the goal in our lab, SEE-Insight. By measuring the distance from the ground truth, these metrics help determine which algorithm provides the most accurate segmentation. 

\vspace{0.5cm}
\noindent
{\em Key Words}: Computer Vision, Image Segmentation, Manual Annotations, Labeled Arrays, Distance Metrics, Fitness Function, Genetic Algorithm.

\end{abstract}
\hrulefill

\section{Introduction}

No single universal segmentation algorithm can address all problems, necessitating tailored selections based on specific tasks. Machine learning (ML) is effective but requires extensive training data, typically limiting its use to large, well-funded projects. Smaller projects often adapt existing algorithms or use manual annotation tools. Deep learning has significantly influenced the development and refinement of segmentation metrics, especially in post-2000 literature.


For a detailed overview of various metrics, including their pros and cons, refer to Table \ref{tab:metrics}. In the table, ``label type" is categorized as either binary or multi-label. Binary masks indicate pixels labeled simply as foreground or background, while multi-label denotes multiple classes. The column ``label invariant" indicates if the metric is sensitive to label naming, and the last column provides the computational complexity with respect to the image size. 

{\footnotesize
\begin{longtable}{|>{\centering\arraybackslash}m{2.5cm}|>{\centering\arraybackslash}m{1.7cm}|>{\centering\arraybackslash}m{3.4cm}|>{\centering\arraybackslash}m{3.8cm}|>{\centering\arraybackslash}m{1.5cm}|>{\centering\arraybackslash}m{1.5cm}|}
\caption{Comparison of different segmentation metrics, including their pros and cons. The columns include ``Label Type" (binary or multi-label), ``Label Invariant" (sensitivity to label naming), and computational ``Complexity." The bolded rows indicate the metrics used in our experiments in this paper.} \label{tab:metrics} \\
\hline
\textbf{Metric Name} & \textbf{Label Type} & \textbf{Pros} & \textbf{Cons} & \textbf{Label Invariant} & \textbf{Complexity} \\
\hline
Jaccard Index (IoU) \cite{Jaccard, choi_survey_2010, IoU} & Binary & Simple to interpret, widely used, suitable for object detection tasks & Less sensitive to smaller objects, does not account for class imbalance & No & $O(N)$ \\
\hline
Dice Coefficient \cite{Dice, VNet} & Binary & Emphasizes overlap, used as a loss function in deep learning models & Less sensitive to smaller objects, does not account for class imbalance & No & $O(N)$ \\
\hline
\textbf{Hamming Distance} \cite{hamming_error_1950} & \textbf{Binary} & \textbf{Computationally efficient} & \textbf{Doesn't account for spatial relationships, ineffective for misaligned masks} & \textbf{No} & \textbf{$\bm{O(N)}$} \\
\hline
Hausdorff Distance \cite{Hausdorff} & Binary & Provides insights into maximum error, useful for precision applications & Computationally intensive, may overemphasize outliers & No & $O(N^2)$ \\
\hline
Matthews Correlation Coefficient (MCC) \cite{Matthews} & Binary & Balanced measure considering true/false positives and negatives & Less intuitive, less commonly used in image segmentation & No & $O(N)$ \\
\hline
Precision-Recall Curve (PR Curve) \cite{PR}  & Binary & Insensitivity to class imbalance & Interpretation complexity, can be misleading if one class is rare, not a single number & No & $O(N)$ \\
\hline
F1 Score, ROC, AUC \cite{sys, ROC, AUC}  & Binary & Insensitivity to class imbalance, single summary metric & Originally for binary classification, threshold dependence & No & $O(N)$ for each threshold \\
\hline
Normalized Cross-Correlation (NCC) \cite{fast, fft} & Binary &  Comparison of spatial alignment and translated or rotated masks  & Computationally intensive, sensitive to noise and intensity variations & No & $O(N. log N)$ if optimized \\
\hline
\textbf{$\bm{\gamma}$-Binary Similarity Measure} \cite{mustafa_modified_2015} & \textbf{Binary} & \textbf{Not sensitive to label assignments}  & \textbf{Not applicable to multi-label arrays} & \textbf{Yes} & $\bm{O(N)}$ \\
\hline
Mean Intersection over Union (mIoU) \cite{miou} & Multi-label & Accounts for class imbalance, per-class and average performance & Averaging might mask poor performance in individual classes & No & $O(N)$ \\
\hline
Fowlkes-Mallows Index (FMI) \cite{FMI} & Multi-label & Extends precision and recall concepts to multi-label scenarios & Can be misleading in imbalanced classes & No & $O(N)$ \\
\hline
Tversky Index \cite{Tversky, tvloss} & Multi-label & Handles imbalanced classes, adjustable sensitivity & Less intuitive, may require tuning of parameters & No & $O(N)$ \\
\hline

\end{longtable}
}

\begin{figure}[h]
\begin{center}
\includegraphics[width=0.8\textwidth]{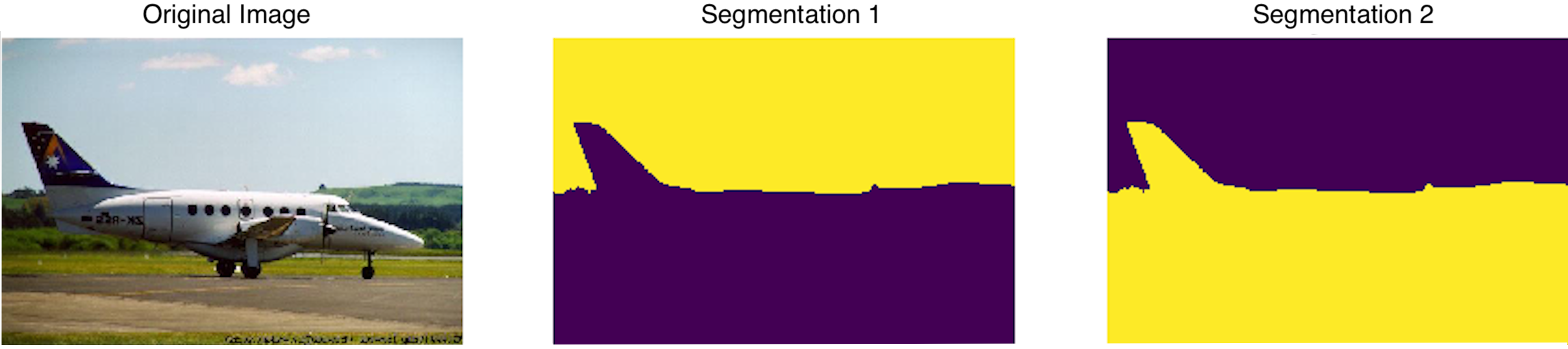}
\caption{Example of binary label segmentation from the Sky dataset \cite{alexandre_ift-slic_2017}, with the original image on the left followed by two segmentations. Both the middle and right images represent identical segmentations of the original, illustrating how a typical labeled array metric evaluates the similarity between the ground truth and the segmented images. Despite different labels being assigned to specific regions in these segmentations, they should ideally be label invariant and yield a distance value of zero.}
\label{fig:different_labels}
\end{center}
\end{figure}

The metrics presented in the table offer a comprehensive toolkit for assessing image segmentation performance. A primary drawback of many of these metrics is that they often require each labeled region to maintain consistent labeling throughout. However, the label names and their ordering are generally unimportant and are assigned based on the mechanics of the individual segmentation algorithm. For instance, consider the images shown in Figure \ref{fig:different_labels}, where different colors represent different labels. Since the labels differ between the two images, the Hamming distance \cite{hamming_error_1950} will register as its maximum (the total number of pixels in the image) even though both segmentations are identical, and the correct distance value between them should be zero. A potential solution is to cycle through the label pairs and return the minimum Hamming distance. However, this method is only feasible for a small number of labels, as the computational effort required grows dramatically with the increase in the number of labeled regions.

A notable binary image comparison metric is the $\gamma$-Binary Similarity Measure, introduced by Mustafa \cite{mustafa_modified_2015}. This similarity metric accurately compares the two binary segmentation images shown in Figure \ref{fig:different_labels} but is not applicable to multi-label arrays. When discussing multi-label segmentation, it is important to recognize the complexity that arises from handling multiple objects or structures in a single image simultaneously.

\begin{figure}[h]
	\begin{center}
		\includegraphics[width=0.6\textwidth]{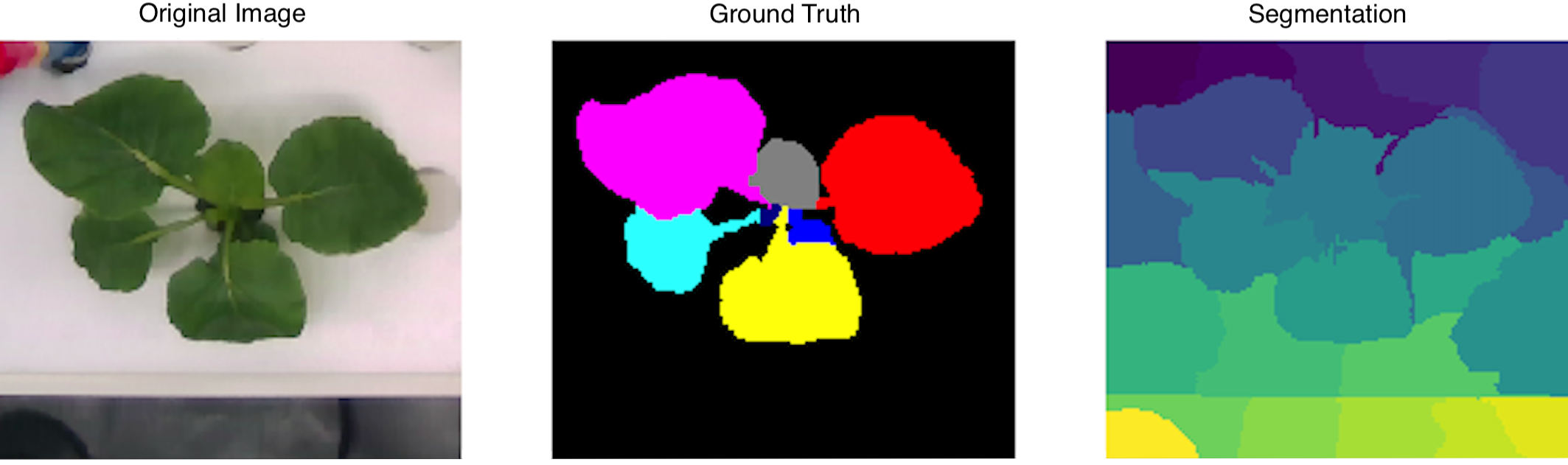}
		\caption{Example of multi-label segmentation of an image. The left image is from the KOMATSUNA dataset \cite{uchiyama_easy--setup_2017}. The middle image is a visual representation of the ground truth provided with the data, with different colors representing different labels. The right image is an example output from a segmentation algorithm attempting to match the ground truth. A typical labeled array metric measures the similarity between the ground truth and the multi-region segmentation.}
        \label{fig:multilabel}
	\end{center}
\end{figure}

A labeled array consists of $N \times M$ pixels with $U$ possible labeled regions, where each pixel is given a unique label $u \in U$ (See Figure \ref{fig:multilabel}). The number of labels within an array can vary, but is natrually bounded between one and the total size of the array ($1 \leq u \leq U \leq N \times M $).

In summary, the intricate nature of multi-label segmentation demands a diverse set of metrics, which have undergone significant refinement and investigation in contemporary research. As with binary segmentation, deep learning has influenced the development of these metrics, but the challenges of handling multiple labels simultaneously introduce a unique set of considerations and requirements.


The goal at SEE-Insight is to develop a tool like SEE-Segment for searching through a space of segmentation algorithms and their hyperparameters via a genetic algorithm to identify the optimal solution for automated segmentation \cite{colbry_toward_2023, see-segment}. For this genetic algorithm, a suitable `fitness function' or segmentation metric that is appropriate for multi-label cases and is insensitive to label permutation or naming is needed. However, as shown in Table \ref{tab:metrics}, none of the metrics, to the best of our knowledge, meet these requirements. The primary contribution of this paper is therefore the creation of such a metric. In the table, the bolded rows indicate the metrics used in the subsequent sections for experiments and comparisons with our developed metrics. Although they do not meet our requirements, we used them in experiments to provide a basis for comparison.

\section{Methodology}

In this paper, we introduce two new labeled array distance metrics, LAD and MADLAD. In this section, we first explain the properties that a metric for evaluating segmentation in multi-label arrays should possess. Then, we describe how we developed LAD and MADLAD as suitable candidates for this purpose.

\subsection{Distance Metric Design Criteria}

The desired metrics are essential for measuring the success of image segmentation on both binary and multi-label arrays. Since successful image segmentation is defined by the problem being solved, the algorithm needs to be compared to some ``ground truth" information regarding the specific segmentation required. The metrics for labeled arrays must: $(1)$ be computationally efficient, $(2)$ provide a distance measurement that can compare the output of multi-labeled arrays, $(3)$ be independent of the naming of the labels, for example, being indifferent to the two segmentations shown in Figure \ref{fig:different_labels}, and $(4)$ use a consistent scale with an understood bias.

\subsection{Design Steps}
\label{sec:methods}

\begin{figure}[h]
	\begin{center}
		\includegraphics[width=0.80\textwidth]{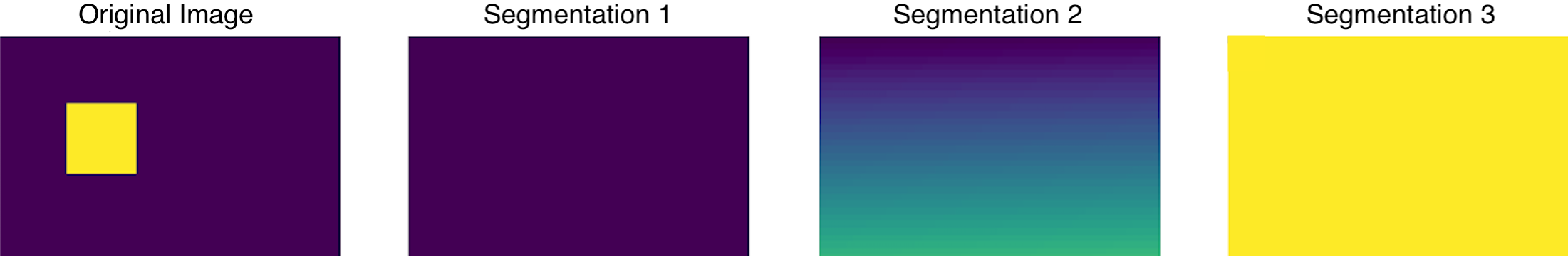}
		\caption{An example of a ground truth image, followed by each of the extreme cases our distance metric needed to account for. The original image acts as the ground truth, the first segmentation is labeled as all background, the second segmentation is multi-label (with different labels for every pixel), and the third segmentation is labeled all foreground.}
        \label{fig:Simple_GT_Example}
	\end{center}
\end{figure}

Consider the simple binary examples shown in Figure \ref{fig:Simple_GT_Example}. In this figure, the ground truth mask has identified a simple box as the foreground and everything else as the background. The three `edge' cases to the right of the mask present extreme segmentation solutions that may need to be compared to the ground truth. In the first segmentation, everything is defined as background, and in the third, as foreground. It is reasonable to assume that any distance metric will produce the same result for both of these cases. The second segmentation acts as an array where each pixel is labeled uniquely, with every pixel given a unique object index. Notice that none of these three cases match the ground truth. Any distance metric used must provide a measure as to how good each result is relative to each other and to the ground truth.   

This section details how we utilized the concepts behind several existing metrics to develop LAD and MADLAD. These steps were instrumental in creating metrics tailored to effectively measure segmentation error. Additionally, subsequent sections will present experiments comparing these metrics with LAD and MADLAD.

\subsubsection{Normalized Hamming Distance (NHD)} 

First, we adapted the Hamming distance \cite{hamming_error_1950} to create the Normalized Hamming Distance (NHD), denoted by $d_{NHD}$. It is designed for binary images, where pixel values are either 0 or 1. The NHD is calculated by first determining the Hamming distance ($d_H$), which counts the number of pixels in the ground truth labeled array ($G$) that do not match the corresponding pixels in the inferred segmentation labeled array ($I$). This discrepancy is then normalized by dividing by the total number of pixels in the image array ($MN$), where $M$ is the number of rows and $N$ is the number of columns in the image as a matrix, resulting in a value between $0$ and $1$. The Hamming distance employs the XOR operation ($\oplus$), identifying mismatches between corresponding pixels of $G$ and $I$. It outputs 0 when both pixels are the same (both 0 or 1) and 1 when they differ. The formulas for $d_{NHD}$ and $d_H$ are given below:
\begin{equation}
\label{eq:Hamming}
d_H = \sum_{i=1}^{M} \sum_{j=1}^{N}(G_{i,j}\oplus I_{i,j}) = \sum_{i=1}^{M} \sum_{j=1}^{N}(G_{i,j} + I_{i,j} - 2 \cdot G_{i,j}I_{i,j}),
\end{equation}
\begin{equation}
\label{eq:Normalized_hamming}
d_{NHD} = \frac{d_H}{MN}.
\end{equation}
In these formulas, $G_{i,j}$ and $I_{i,j}$ refer to each pixel of $G$ and $I$ indexed by $i$ and $j$ as rows and columns.

An NHD of 0 indicates no discrepancy between the ground truth and the inferred segmentation, signifying identical segmentations. Conversely, an NHD of 1 implies no commonality between the segmentations, where no pixels match between the ground truth and the inferred segmentation.

While NHD provides a straightforward measure of discrepancy, it falls short in scenarios where segmentations are identical in structure but differ only in their labeling. For example, two segmentations that are identical except for the labels assigned to regions (like in Figure \ref{fig:different_labels}) would result in a high NHD value, suggesting a high level of discrepancy when, functionally, the segmentations are the same. This limitation highlighted the need for a metric capable of recognizing equivalent segmentations irrespective of label differences.

\subsubsection{Binary Similarity Measure (BSM)}

To address this limitation, we use the Binary Similarity Measure (BSM) from \cite{mustafa_modified_2015}, a metric specifically tailored to evaluate binary labeled images. This is crucial for accurately assessing binary segmentation results, where traditional metrics may fail to account for variations in labeling conventions.

BSM uniquely addresses discrepancies when identical segmentations are labeled differently across datasets. It is designed for binary images, where pixels are either 0 or 1 (or segmentations that have only `foreground' and `background' labels). Here, $M$ represents the number of rows, $N$ the number of columns in the image matrix, and $d_H$ is the Hamming distance as previously explained. The core of BSM's functionality is expressed as $\left| 1 - \frac{2 \cdot d_H}{MN} \right|$ or $\left| 1 - 2 \cdot d_{NHD} \right|$, according to \cite{mustafa_modified_2015}.

BSM provides a comprehensive indicator of how similar two binary images are, effectively measuring how well the inferred segmentation aligns with the ground truth. BSM calculates the overall similarity by assessing the entire image and determining the absolute deviation from 1 of twice NHD (or the average dissimilarity across all pixel pairs). It captures the similarity between corresponding pixels, regardless of the label values assigned. Since the original BSM from \cite{mustafa_modified_2015} captures similarity and not the difference, we adapted it by subtracting it from 1 to create a distance metric, denoted by $d_{BSM}$, to make it comparable with our other metrics. Therefore, we have:
\begin{equation}
\label{eq:gamma}
d_{BSM} = 1 - \left| 1 - \frac{2 \cdot d_H}{MN} \right| = 1 - \left| 1 - 2 \cdot d_{NHD} \right|.
\end{equation}
When two images are identical, their $d_H$ is zero, and $d_{BSM}$ becomes zero. In cases where two segmentations are identical but with completely different labels (e.g., swapped 0 and 1 labels), all pixels are counted as ``differing" so $d_H$ equals $MN$, the total number of pixels, and $d_{BSM}$ again becomes zero, indicating a perfect match as intended. In contrast, a simple or normalized Hamming distance would report the latter case as an extremely poor segmentation.

While $d_{BSM}$ effectively addresses binary labeling discrepancies, it does not accommodate multi-label segmentations. Its application is limited to binary cases, and thus it cannot handle segmentations with more than two label types. Recognizing this limitation led to the development of further metrics.

\subsubsection{Region Mapping (RM)} 

Initially, we developed the Region Mapping (RM) formula to handle multiple labels, evaluating the correspondence between regions across two labeled images. In this approach, an inferred labeled array $I$ with $V$ regions and a ground truth labeled array $G$ with $U$ regions are considered. The RM formula determines which region in $I$ has the largest overlap with a region in $G$, assigning the region in $I$ the corresponding ground truth label $g$, where $1 \leq g \leq U$. Pixels within each region in $I$ that do not map onto their designated $g$ are counted and added to a total mismatch score $P$. This mapping process assesses how effectively each label in the inferred labeled array correlates with those in the ground truth, quantified by $P$, the total number of pixels that do not align properly under this mapping.

This mapping process is then used to define the distance metric $d_{RM}$, which is the proportion of misaligned pixels relative to the total number of pixels in the image, providing a measure of segmentation error:

\begin{equation}
\label{eq:RegionMapping}
d_{RM} = \frac{P}{MN}.
\end{equation}
The value of $d_{RM}$ ranges between 0 and 1, where 0 indicates a perfect match with no mismatches, and 1 indicates the worst case with no matches at all.

However, we encountered the following challenges with Region Mapping:

\noindent\textbf{Directionality:} RM considers the order of comparison (from $I$ to $G$). When one of the images is a known ground truth, this directionality ensures that the calculation accurately assesses how well the inferred segmentation maps onto the ground truth.

\noindent\textbf{Over-segmentation:} RM does not account for the number of regions within the segmentation. This omission can yield misleading results in scenarios of over-segmentation. As illustrated in the multi-label column of Table \ref{tab:MADLAD:1}, an inferred segmentation that is highly fragmented may inaccurately appear highly similar to a less fragmented ground truth, provided that the largest fragments overlap significantly.

\subsubsection{Labeled Array Distance (LAD)} 

We developed the Labeled Array Distance (LAD) to address the issue of over-segmentation by considering the total number of labels in both the ground truth and the comparison labeled array.

The LAD distance metric, denoted by $d_{LAD}$, builds upon the region mapping approach by incorporating both the difference in the number of regions and the total number of mislabeled pixels for a given mapping. The sum of these two quantities is then normalized by the total number of image pixels. If two images match perfectly, $d_{LAD}$ is 0, while no agreement results in a high value with a maximum of 1. 

The formula for LAD is as follows:
\begin{equation}
\label{eq:LAD}
d_{LAD} = \frac{P + |U - V|}{MN}.
\end{equation}
Here, $U$ represents the total number of labels in the ground truth array ($G$), $V$ is the total number of labels in the inferred labeled array ($I$), and $P$ is the total number of pixels that do not match under the `best' mapping between the arrays, similar to the RM metric. This method effectively quantifies the overall discrepancy, taking into account both label mismatches and spatial inaccuracies.

\subsubsection{Mismatch Adjusted Difference for the Labeled Array Distance (MADLAD)} 

In the basic LAD model, the factors of pixel error and region number mismatch are given equal weight. However, as indicated by the data in Table \ref{tab:MADLAD:1} and often observed in practical scenarios, the mismatch in the number of regions can be more significant than the mere count of misaligned pixels. To address this, we designed the Mismatch Adjusted Difference for the Labeled Array Distance (MADLAD) metric, which introduces an exponential adjustment to more heavily weight the region mismatch.


The MADLAD formula modifies the weighting of region mismatches by employing an exponential factor that significantly influences the overall distance calculation, particularly when discrepancies in region counts are notable. It is denoted by $d_{MADLAD}$ and defined as follows:
\begin{equation} 
\label{eq:MADLAD}
d_{MADLAD} = \left(\frac{P}{MN} + \frac{|U-V|}{U+V}\right)^{\left(1 - \frac{|U-V|}{U+V}\right)}.
\end{equation}
Here, $P$, $U$, and $V$ are represented as in LAD and RM. The base of the exponent combines the pixel error, normalized by the total number of image pixels ($MN$), with the proportional difference in the number of labels. The exponent itself $\left(1 - \frac{|U-V|}{U+V}\right)$ reduces the overall impact of this base value as the mismatch in label counts increases. Essentially, when the label mismatch is small, the power applied is closer to 1, having less of an effect on reducing the base value. However, as the mismatch increases, the power decreases towards zero, causing the computed $d_{MADLAD}$ value to approach 1, indicating a significant mismatch. This design ensures that MADLAD becomes especially sensitive to errors in scenarios where there is a large discrepancy in the number of regions, reflecting the critical importance of having similar region counts for accurate segmentation.

Table~\ref{tab:MADLAD:1} effectively summarizes how different metrics respond to extreme segmentation cases illustrated in Figure \ref{fig:Simple_GT_Example}. The metrics vary significantly in their evaluations of background-only, multi-label, and foreground-only segmentations. The ``Same" column in the table represents a scenario where an image is compared with itself, where logically, we expect the distance to be zero as there are no differences between the images. 

It's important to clarify a few points to improve the understanding of the results and the behaviors of these metrics:

\noindent\textbf{1. NHD:} Reflects changes dramatically with the order of labeling due to its sensitivity to label mismatches. It shows a high distance for multi-label scenarios, indicating a substantial mismatch.

\noindent\textbf{2. BSM:} This metric is less sensitive to label mismatches in binary scenarios, showing low variation across different types of segmentation errors.

\noindent\textbf{3. RM:} Appears highly effective in some cases due to its focus on the largest matching regions. It shows no distance in the multi-label scenario, suggesting perfect overlap, which may not realistically reflect the error in typical use cases.

\noindent\textbf{4. LAD:} Shows consistency in distinguishing the extremity of segmentation differences but rates the multi-label scenario as completely erroneous (distance of 1), highlighting its sensitivity to label number mismatches.

\noindent\textbf{5. MADLAD:} Similar to LAD in its responses, with adjustments that make it sensitive to the scale of mismatches in labeling, also rating the multi-label scenario as maximally distant.

\begin{table}[h!]
    \centering
    \caption{Comparison of different metrics applied to labeled arrays, as illustrated in Figure \ref{fig:Simple_GT_Example}. The metrics are evaluated for cases where images are segmented as all background, multi-label, and all foreground. The ``Same" column refers to scenarios where an image is compared with itself, where a distance of zero is expected.}
    \label{tab:MADLAD:1}
    \footnotesize
    \begin{tabular}{|l|c|c|c|c|}
        \hline
        \textbf{Metric} & \textbf{Same} & \textbf{Background} & \textbf{Multi-label} & \textbf{Foreground} \\
        \hline 
        \textbf{NHD} & 0.00 & 0.95 & 0.99 & 0.04 \\
        \hline 
        \textbf{BSM} & 0.00 & 0.08 & 0.08 & 0.08 \\
        \hline 
        \textbf{RM} & 0.00 & 0.04 & 0.00 & 0.04 \\
        \hline 
        \textbf{LAD} & 0.00 & 0.04 & 1.0 & 0.04 \\
        \hline 
        \textbf{MADLAD} & 0.00 & 0.04 & 1.0 & 0.04 \\
        \hline
    \end{tabular}
\end{table}

The table shows that while all metrics can effectively detect when two arrays are identical (distance of zero), only NHD, LAD, and MADLAD provide meaningful assessments for more complex multi-label arrays. NHD’s results show that its sensitivity to the order of labeling can lead to variable outcomes, indicating that while it can handle complex arrays, its reliability might be compromised without consistent labeling. Both LAD and MADLAD offer a robust evaluation of segmentations, particularly in recognizing the severity of mismatches in more complicated label configurations.

\section{Experiments}

This section reviews a series of experiments conducted to demonstrate the utility of the LAD and MADLAD distance metrics in multi-label segmentation, comparing them to the BSM and NHD segmentation metrics introduced in Section \ref{sec:methods}. The first experiment (Section \ref{sec:Binary_Comparison}) involves manipulating a binary segmentation image by applying noise or common morphological operations to mimic segmentation errors. This experiment aims to show that LAD and MADLAD yield similar, and often identical, results in these basic scenarios. The second experiment in Section \ref{sec:GT_fitness} involves multiple ``ground truth" examples from challenging, real-world images to establish a reasonable benchmark and identify a cut-off threshold for distance metric values when using LAD and MADLAD on binary images. The third experiment (Section \ref{sec:which_is_better}) assesses the effectiveness of these distance metrics on real-world images by measuring their alignment with human judgment using a methodology based on human preference assessments. Finally, the last experiment in Section \ref{sec:example_app} illustrates the application of LAD and MADLAD as primary metrics in a search algorithm aimed at finding reasonable segmentations for real-world problems.  

\subsection{Evaluation Through Image Manipulations}
\label{sec:Binary_Comparison}
This section compares the distance metrics across five different scenarios for simulating errors: salt noise, pepper noise, salt-and-pepper noise, closing operation, and opening operation. These experiments focus on binary comparisons where a 0 label (black) is considered foreground and a 1 label (white) is considered background. 

\subsubsection{Salt Noise}
In this initial experiment, salt noise is applied to the segmentation image of a chameleon to observe how various distance metrics react to random noise within a bounded background region. As depicted in Figure \ref{fig:sp}–left, the horizontal axis ranges from 0\% to 100\%, where 0\% indicates no salt noise added, and 100\% indicates that the image is fully composed of salt noise (all pixels are salt noise, making the image entirely white). 

The results reveal a linearly increasing trend across all distance metrics. However, an outlier occurs with MADLAD when it encounters a degenerate case, where all labels in one array are mapped to a single label in the target array. Typically, all metrics, including MADLAD, scale between 0 and 1 under normal circumstances. However, to clearly mark this degenerate scenario, the MADLAD value is explicitly and manually set to 1.5 using conditional statements in the code. Results for LAD and MADLAD overlap with the NHD metric, except in the degenerate case. In the plot, the BSM values are observed to be twice the NHD values, consistent with Equation \ref{eq:gamma}. According to Equation \ref{eq:gamma}, when the value of $d_{NHD}$ is not greater than $0.5$ (as is the case here), $d_{BSM}$ is equal to $2\cdot d_{NHD}$.

\subsubsection{Pepper Noise}

\begin{figure}[h]
\centering
		\includegraphics[width=1\textwidth]{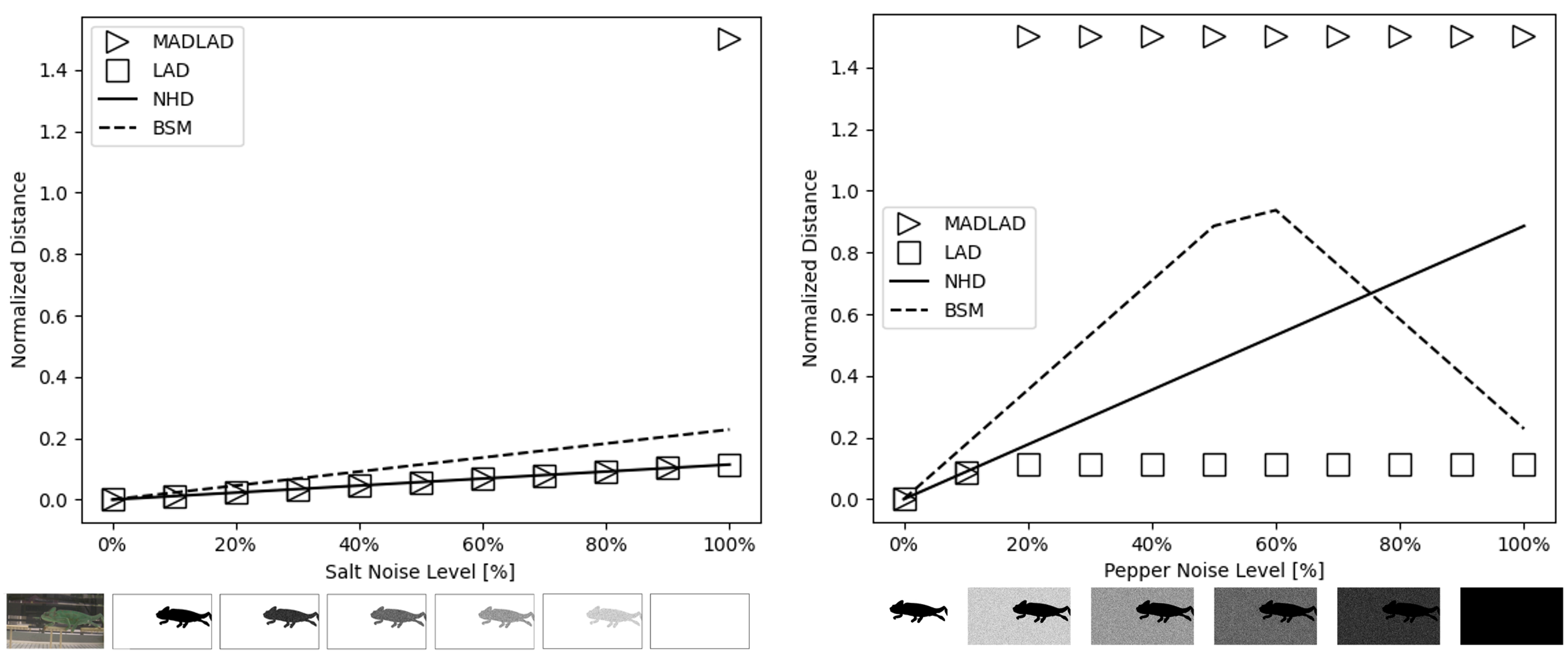}
		\caption{Comparison of four example metrics for the salt noise scenario (left) and pepper noise scenario (right). Below the plot, the original chameleon image is shown at the far left, followed by segmentation masks with progressively increasing noise.}
        \label{fig:sp}
\end{figure}

As depicted in Figure \ref{fig:sp}–right, pepper noise differs from salt noise primarily in the size of the regions affected by the noise. Given that the background in this particular chameleon image is larger than the foreground, the degenerate case (where both labels are mapped to the same label) for the MADLAD formula occurs earlier.

LAD, in contrast to MADLAD, does not recognize an early degenerate case because we did not add a conditional statement for LAD to mark the degenerate case by setting the value to 1.5; this was implemented only for MADLAD. Consequently, LAD maintains a consistent error level since both regions are mapped to the same label. 

Similar to the salt noise case depicted in Figure \ref{fig:sp}–left, BSM is twice the NHD when NHD is small. However, BSM fails to detect the degenerate case and responds as if there were a label swap, beginning to decrease–a typical response observed in less sophisticated metrics. Additionally, BSM is not symmetric due to the ratio of the object to the image. The object is smaller than the background, causing the peak to shift to the right rather than being centered, and the background's larger size influences the noise. NHD still exhibits the previous linearly increasing trend. Furthermore, when the pepper noise is at its maximum or 100\%, the value of NHD is approximately $0.9$. In contrast, with 100\% salt noise, NHD is around $0.1$. This difference is consistent with the ratio of the size of the chameleon to the background, which is approximately 10\% to 90\%.

  \begin{wrapfigure}{R}{0.45\textwidth}
	\begin{center}
        \includegraphics[width=0.45\textwidth]{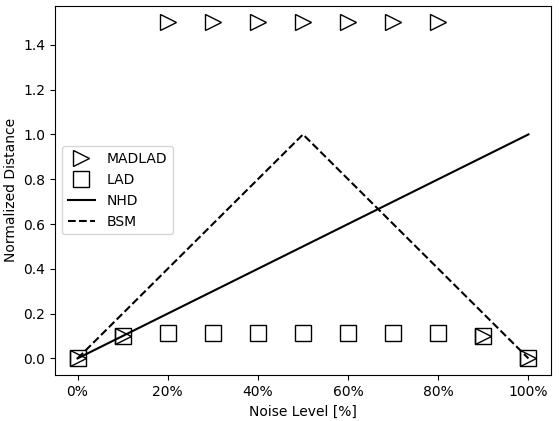}
        \includegraphics[width=0.45\textwidth]{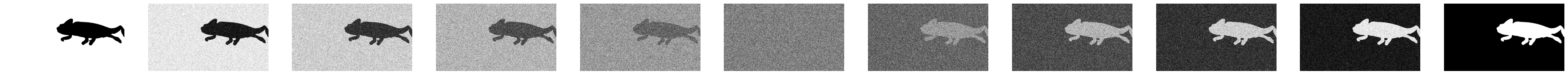}
     	\caption{Combination of salt and pepper noise.  This graph demonstrates that the NHD metric fails to recognize a swap in labels, whereas and MADLAD encounter a degenerate case at mid-level noise.  BSM performs optimally in this scenario but is constrained to binary labels only.}
        \label{fig:salt_and_paper}
	\end{center}
\end{wrapfigure} 

\subsubsection{Salt-and-Pepper Noise}
This section explores the effects of combining salt and pepper noise at each step. After 10 iterations, 100\% noise has been applied, fully saturating the base segmentation. The noise is applied separately to the background and foreground, resulting in a complete label swap. This means that all pixels are inverted relative to their initial state, as illustrated in Figure \ref{fig:salt_and_paper}.

The NHD metric fails to capture this label swap because it strictly evaluates the matching of specific labels without considering their spatial inversion. Consequently, it reaches the maximum value of 1 at the 100\% noise level. Conversely, the distance metrics LAD, MADLAD, and BSM successfully detect this behavior. Notably, similar to Figure \ref{fig:sp}–right, MADLAD reaches a degenerate case early, where both labels are mapped to the majority label. Without this degenerate case, LAD and MADLAD would yield similar outcomes in the binary case. Due to the gradual addition of noise and the complete inversion of all pixel values, all measures except NHD exhibit symmetry, with BSM peaking in the middle. Once again, BSM is twice NHD when NHD is small.

\subsubsection{Closing and Opening Operations}

Although salt-and-pepper noise examples provide insightful demonstrations of the differences between metrics, they typically do not represent the kinds of errors commonly seen in segmentation tasks. In the upcoming experiments, we will explore more realistic types of segmentation errors using binary morphological operations, starting with the closing operation.

\begin{figure}[h]
	\begin{center}
        \includegraphics[width=1\textwidth]{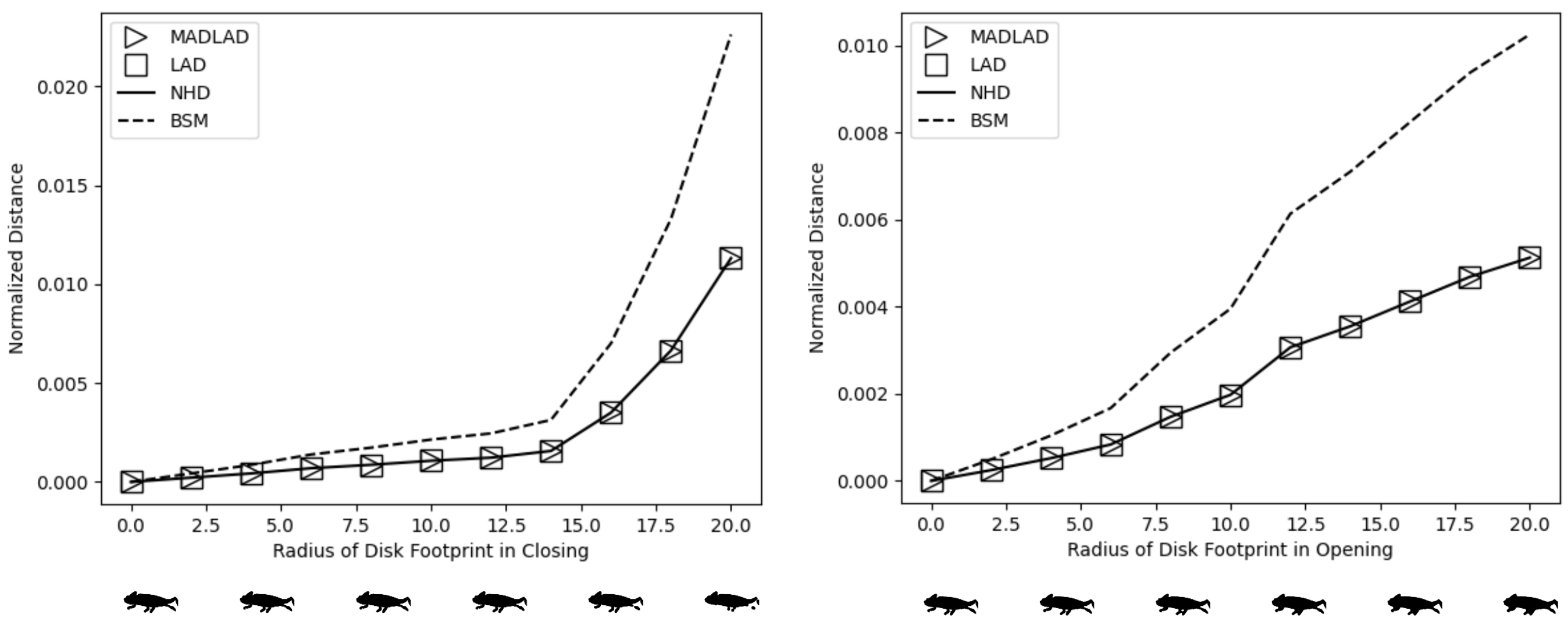}
     	\caption{Iterative application of increasingly larger closing operations (left) and opening operations (right) on the chameleon image. Notice that all the metrics yield similar outcomes, with LAD and MADLAD results closely aligning with those of the NHD metric. However, the BSM metric shows slightly higher values.}
        \label{fig:openclose}
	\end{center}
\end{figure}

The closing operation in binary morphology involves a sequence of binary dilation followed by binary erosion, resulting in more realistic alterations to the boundaries of the shapes. With each iteration, the footprint of the closing operations is increased, as shown in Figure \ref{fig:openclose} at the bottom of the plot for the chameleon's segmentation images. Conversely, the opening operation consists of an erosion followed by dilation. Like the closing operation, the size of the footprint increases with each iteration, realistically altering the shape boundaries and mimicking errors commonly encountered in real-world segmentation tasks.

Both operations result in slight changes in the values from the distance metrics due to subtle adjustments on the boundary between the background and foreground, as visible in the plot in Figure \ref{fig:openclose}. It can be observed that LAD, MADLAD, and NHD are all equal to each other. Additionally, since NHD is always less than 0.5 in these cases, the value of BSM is always twice the NHD, according to Equation \ref{eq:gamma}.

\subsection{Evaluating Manual Segmentation Variability}
\label{sec:GT_fitness}
The second set of experiments employs LAD and MADLAD to compare ground truth results obtained from different users. The primary objective of this experiment is to illustrate the inherent challenges of segmenting real-world images and to demonstrate that there is no definitive ``ground truth" for any given problem. These results are used to establish reasonable thresholds for the LAD and MADLAD metrics. These thresholds are crucial for defining acceptable levels of segmentation error in the example application detailed in Section \ref{sec:example_app}, helping to determine when a segmentation result is sufficiently close to these varying ground truths.

\begin{figure}[h]
	\begin{center}
		\includegraphics[width=0.6\textwidth]{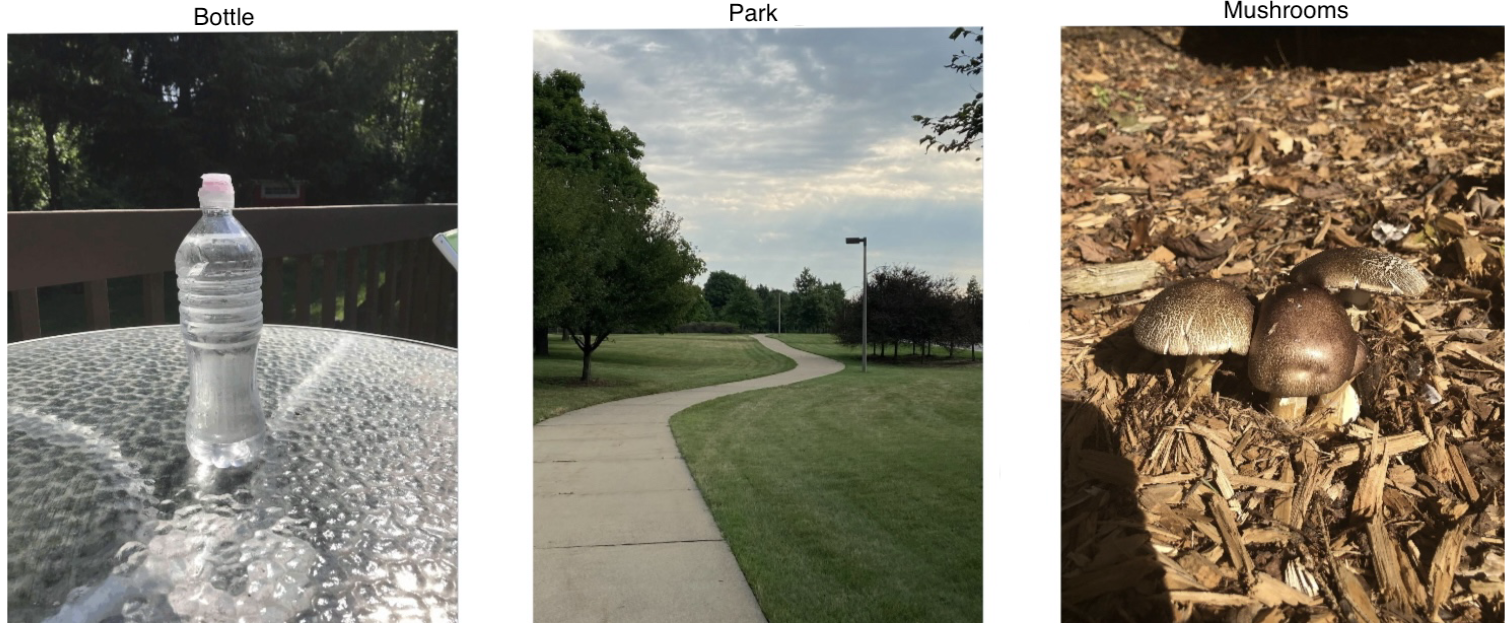}
		\caption{Set of three images to be segmented by experiment participants. Taken from the SEE-Insight Example Images repository \cite{github-see-segment}.}
        \label{fig:imageset}
	\end{center}
\end{figure}

This experiment involves a set of three images, shown in Figure \ref{fig:imageset}, each segmented by six participants who vary in their experience with manual segmentation. To ensure consistency among participants, comprehensive instructions were provided on how to install and use the GNU Image Manipulation Program (GIMP) \cite{gimp} for annotating images. 

The collected images showcase the variability typical in manual segmentation, serving as ground truths. Within these images, color values are assigned such that zero represents black and one represents white. 

\begin{figure}[h]
	\begin{center}
		\includegraphics[width=0.7\textwidth]{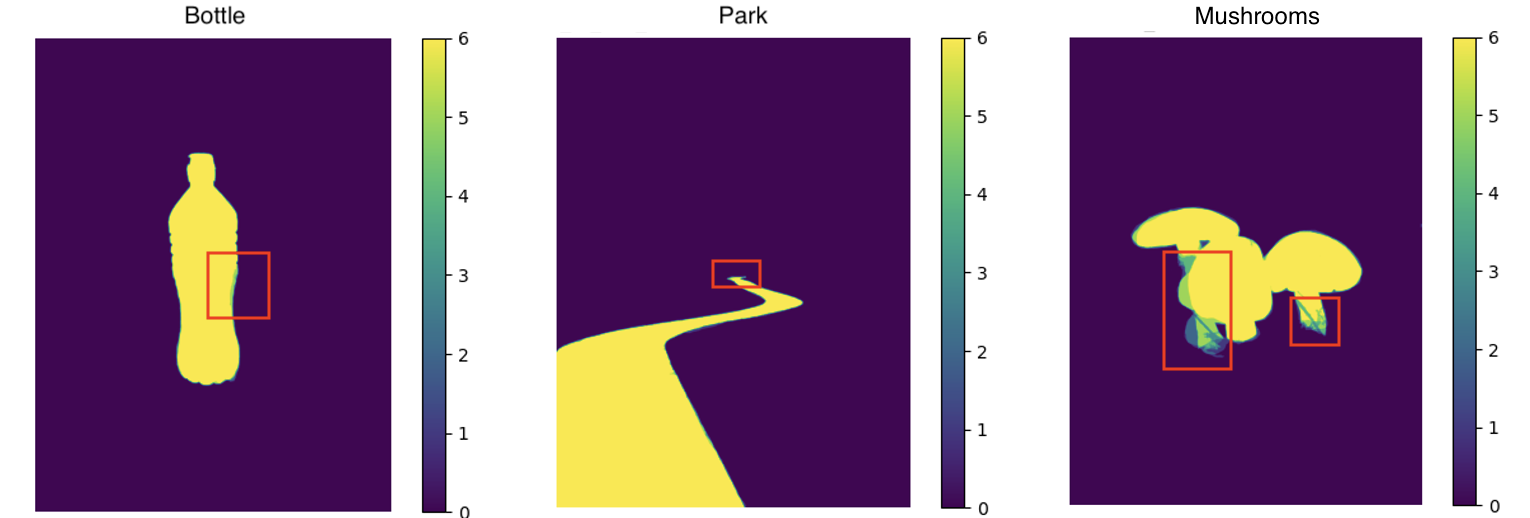}
      \caption{Composite images derived from summing pixel color values across six participant-segmented images for all three scenes shown in Fig. \ref{fig:imageset}.  Values of zero (dark blue) and six (dark red) indicate unanimous agreement among participants. Values between zero and six signify varying degrees of disagreement. Areas of significant disagreement among participants are highlighted with red boxes in each image.}
    	\label{fig:fitcompare}
	\end{center}
\end{figure}

These images are overlaid, and the pixel values are aggregated to create `sum images', as illustrated in Figure \ref{fig:fitcompare}. A pixel value of six at any given point indicates that all participants used white for segmentation at that location. Conversely, a value of zero indicates that all used black. Values between zero and six denote areas of disagreement among the participants. For example, if five out of six participants labeled a spot as black and one as white, the resulting value would be one, effectively highlighting the variation between segmentations.

Figure \ref{fig:fitcompare} displays a comparison of segmented images for each of the three photos. Areas of significant disagreement among participants are highlighted with red boxes in each image.

These findings underscore the complexity involved in defining `ground truth' in image segmentation. There is no universal standard of ground truth against which machine-segmented images can be consistently measured. However, manual segmentation data provide a solid baseline for evaluating the accuracy of segmentation results. In Table \ref{tab:waterbottle:1}, the degree of agreement among different manual segmentations is quantified for the 'Bottle' image, with zero representing a perfect match.

\begin{table}[h!]
    \centering
    \caption{Comparative analysis using the LAD metric for manually segmented `Bottle' images. Values less than 0.0015 are highlighted in bold, indicating a very good fit.}
    \label{tab:waterbottle:1}
    \footnotesize
    \begin{tabular}{ |c|c|c|c|c|c|c| }
        \hline
        \textbf{\footnotesize Participant} &  \textbf{1} &  \textbf{2} &  \textbf{3} &  \textbf{4} &  \textbf{5} &   \textbf{6} \\
        \hline 
         \textbf{1} & \textbf{0.0000} & 0.0025 & 0.0023 & 0.0022 & 0.0023 & 0.0018 \\
        \hline 
         \textbf{2} & 0.0025 & \textbf{0.0000} & \textbf{0.0011} & \textbf{0.0011} & 0.0018 & 0.0015 \\
        \hline 
         \textbf{3} & 0.0023 & \textbf{0.0011} & \textbf{0.0000} & \textbf{0.0010} & 0.0015 & \textbf{0.0013} \\
        \hline 
         \textbf{4} & 0.0022 & \textbf{0.0011} & \textbf{0.0010} & \textbf{0.0000} & \textbf{0.0014} & \textbf{0.0012}  \\
        \hline 
         \textbf{5} & 0.0023 & 0.0018 & 0.0015 & \textbf{0.0014} & \textbf{0.0000} & \textbf{0.0013} \\
        \hline
         \textbf{6} & 0.0018 & 0.0015 & \textbf{0.0013} & \textbf{0.0012} & \textbf{0.0013} & \textbf{0.0000}\\
        \hline
    \end{tabular}
\end{table}

The table is symmetric, as expected, because the distance metric is independent of the order of comparison. For instance, the metric value when Participant 1 is compared to Participant 2 is identical to when Participant 2 is compared to Participant 1. Furthermore, all entries along the main diagonal are zero, indicating a perfect match when participants compare their own segmentations against themselves. The tables for `Park' and `Mushrooms' were similar, so we only included the `Bottle' table for brevity.

In the evaluation of manual segmentations, differences in consistency and accuracy among various images were noted. Excluding the main diagonal entries, which compare images to themselves, variations were observed in the consistency of segmentation among different images:

\noindent\textbf{1. Bottle:} Out of the 30 non-diagonal cells, 14 displayed a distance metric value less than 0.0015, indicating a very close match. The average distance metric for these cells was 0.0013, showcasing the highest consistency in segmentation among participants for this image set.

\noindent\textbf{2. Park:} This image set showed slightly lower consistency, with 12 out of the 30 cells having a distance metric less than 0.0015. The average value for these cells was 0.0017, suggesting a decent, but slightly less consistent, segmentation quality compared to the `Bottle' image.

\noindent\textbf{3. Mushrooms:} This set did not have any cells with a distance metric under 0.0015. The average value across the 30 non-diagonal cells was 0.0089, indicating the lowest consistency. The busy background of this image likely made it difficult for participants to consistently determine the edges of the subject, leading to the highest variation in segmentation results.

These findings underscore the impact of image complexity on manual segmentation tasks. Images with clearer and less cluttered backgrounds, like `Bottle', tend to yield more consistent segmentations, whereas those with busier or similarly-colored backgrounds, such as `Mushrooms', present greater challenges in achieving uniform segmentation outcomes.

Ultimately, manual segmentation is not only time-consuming but also yields variable ground truths that depend on the individual user and the specific image in question. The collected data provides a baseline for evaluating the consistency between manually segmented images, which can be utilized to establish distance metric requirements for automated segmentation tools.

\subsection{Evaluating Segmentation Metrics Against Human Judgment}
\label{sec:which_is_better}
This section evaluates the effectiveness of four segmentation metrics—LAD, MADLAD, BSM, and NHD—as introduced in Section \ref{sec:methods}, in reflecting human judgments of segmentation closeness. Since all segmentation metrics inherently exhibit biases that can influence their outcomes, identifying a metric that accurately mirrors human evaluative processes is crucial. To achieve this, our experiment employs the Elo rating system \cite{elo}, a method traditionally used in chess to rank competitors based on pairwise comparisons.

We conducted an experiment using image segmentations of three photos displayed in Figure \ref{fig:imageset}. Team members served as evaluators, rating these segmentations to determine which metrics most closely resemble human judgment. A linear regression analysis was utilized to compare the distances derived from the metrics with those derived from human ratings, aiming to pinpoint the metric that best approximates human perceptual accuracy.

The experimental platform was developed in Python, using the Tkinter library \cite{tkinter} for graphical user interface (GUI) creation and the Python Imaging Library (PIL) \cite{pillow} for image manipulation. The setup was designed to facilitate the direct comparison of image segmentations by human participants.

Images were categorized into three distinct scenes: a park, mushrooms, and a water bottle. For each category, original images and their segmentations were stored in separate folders. A specific function excluded original images from the selection, loading only segmentations for comparison. Each segmentation was then paired with every other segmentation within its category to form unique evaluation pairs.

\begin{wrapfigure}{R}{0.3\textwidth}
    \centering
    \includegraphics{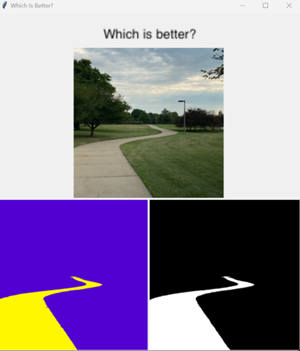}
    \caption{Experimental GUI setup: This interface displays the original image at the top for reference, with pairs of segmented images shown side by side below. Participants select the segmented image they perceive as most closely matching the original image.}
    \label{fig:WIBgui}
\end{wrapfigure}

The GUI, as shown in Figure \ref{fig:WIBgui}, displayed the original image at the top of the window for reference, with pairs of segmented images presented side by side below it. Participants indicated their preference by clicking on the segmented image they believed most closely matched the original image. Each participant's selection within the image pairs was recorded, incrementally building a choice matrix. This matrix represented the collective judgments of the participants across all image pairs within each category and it was exported to a CSV file at the end of the experiment. 

The Elo rating system was adapted to evaluate image segmentation quality. This system normalized the data and provided a comparative rating for each segmented image based on pairwise comparisons. Each image started with an initial rating of zero, which was adjusted according to the outcomes of the pairwise comparisons. Adjustments were made using a constant factor ($K=32$), resulting in a final set of Elo ratings for each image that reflected its relative quality as judged by the participants.

Additionally, distance matrices for each image category were calculated using the four metrics previously mentioned: LAD, MADLAD, BSM, and NHD. Each of these metrics is designed to quantitatively assess segmentation quality by quantifying the dissimilarity between every pair of segmented images.

To determine the alignment between human judgments (as represented by Elo rating-derived distances) and quantitative assessments (as represented by metric-derived distances), linear regression analyses were conducted. These analyses evaluated the correlation strength and significance between the two sets of distances, thereby identifying which metrics most closely approximated human evaluation of the segmentations.

\begin{figure}[h]
    \centering
    \includegraphics[width=0.6\linewidth]{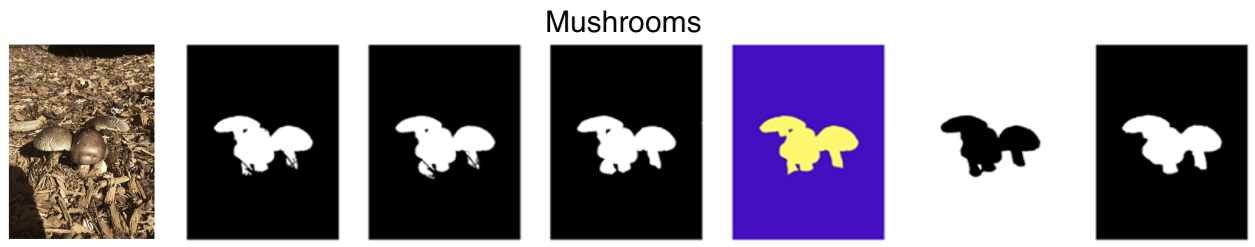}
    \caption{The `Mushrooms' image at far left and the ranking of its segmentations by quality. Segmentations are displayed from best (left) to worst (right) based on human evaluations. Note that the different colors represent different labels used by different people performing the segmentation.}
    \label{fig:ranking}
\end{figure}

Figure \ref{fig:ranking} displays the original `Mushrooms' image along with the rankings of its segmentation image quality as determined by human participants using the Elo rating system. In the figure, the segmentation positioned next to the original image is the ``winner of the Elo tournament," indicating it emerged as the highest ranked based on human evaluations and is considered the best segmentation. Conversely, the segmentations on the far right are ranked as the worst, with all images systematically ordered from best to worst moving left to right.

An important observation is that there are more immediately distinguishable differences between the segmentation sets for the `Park' and `Mushroom' images, whereas the differences between the `Bottle' segmentations are more subtle. This subtlety could influence the relationship between human perceptions and quantitative evaluations.

The linear regression analysis conducted on the distance measurements derived from both human judgments (Elo ratings) and the four distance metrics (MADLAD, LAD, BSM, and NHD) revealed varying degrees of correlation across the three images: `Park', `Mushrooms', and `Bottle'. A stronger and more statistically significant positive correlation between the metric-derived distances and the Elo-derived distances indicated a closer alignment with human judgment.

\begin{figure}[h]
    \centering
    \includegraphics[width=1\linewidth]{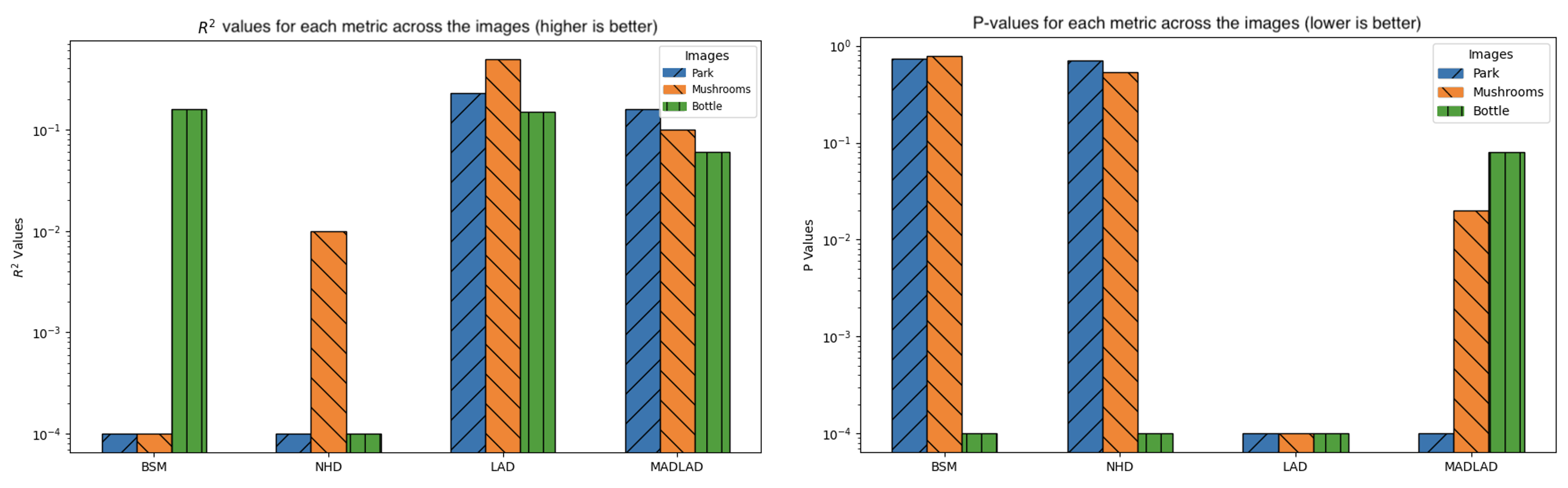}
    \caption{The $R^2$ and $P$-values for each distance metric across the three image categories. Higher $R^2$ values indicate better correlation with human judgments, showing stronger alignment with human perception. Lower $P$-values suggest statistically significant correlations, indicating the results are not due to chance. A logarithmic scale is used on the y-axis for better visibility.}
    \label{fig:R2P}
\end{figure}

In regression analysis, the $R^2$ value, or coefficient of determination, quantifies the proportion of the variance in the dependent variable that is predictable from the independent variables. Ranging from 0 to 1, a higher $R^2$ value indicates a better fit, suggesting that the model explains a greater proportion of variance. It provides an indication of goodness of fit and, therefore, a measure of how well unseen samples are likely to be predicted by the model. The $p-$value, on the other hand, tests the null hypothesis that the coefficient is zero (no effect). A low $p-$value $(< 0.05)$ indicates that we can reject the null hypothesis, typically suggesting the model findings are statistically significant and not due to chance. Conversely, a higher $p-$value suggests less confidence in the model's predictive power regarding the specific variable's influence. The $R^2$ values and corresponding $p-$values for each of the four metrics with respect to each of the three images are displayed in Figure \ref{fig:R2P}.
 
For `Park', MADLAD distances exhibited a slight but statistically significant positive correlation with Elo distances, indicated by an $R^2$ of approximately $0.16$ and a $p-$value of $0.004$. This suggests a real, albeit weak, relationship where changes in MADLAD distances somewhat predict changes in Elo ratings. LAD distances displayed a somewhat stronger positive correlation with Elo distances, with an $R^2$ of approximately $0.23$ and a $p-$value of approximately $0.0004$, indicating a higher correlation with human judgment. The correlations between distances from BSM and NHD and Elo distances were weaker, with both metrics yielding an $R^2$ of approximately $0$ and $p-$values of $0.73$ and $0.70$, respectively, suggesting no significant relationship.

For `Mushrooms', MADLAD distances showed a weak but statistically significant correlation with Elo distances, with an $R^2$ of $0.10$ and a $p-$value of approximately $0.02$. This indicates a slight predictive relationship where MADLAD distances are somewhat indicative of Elo ratings. LAD distances, however, demonstrated a considerably stronger correlation with Elo distances, evidenced by an $R^2$ of $0.49$ and a $p-$value of approximately $0$, reflecting a more robust alignment with human judgments. Similar, the correlations for NHD and BSM were weaker, with $R^2$ values of $0.01$ and $0$ and $p-$values of approximately $0.53$ and $0.78$ respectively, indicating negligible relationships.

For `Bottle', MADLAD distances demonstrated a weak positive correlation with an $R^2$ of $0.06$ and a $p-$value of approximately $0.08$, suggesting a near-significant relationship where changes in MADLAD distances slightly predict changes in Elo ratings. LAD showed a higher correlation with human judgment, evidenced by an $R^2$ of $0.15$ and a significant $p-$value of approximately $0.004$. NHD continued to show a negligible correlation ($R^2$ of 0), while BSM displayed a slight positive correlation with an $R^2$ of $0.16$ and a significant $p-$value of approximately $0$, suggesting a slight predictive relationship with human ratings.

Overall, the LAD distance metric demonstrated the most consistent positive and statistically significant correlation across all three images, indicating the best correlation with human judgment, followed by MADLAD. In contrast, neither NHD nor BSM distances consistently correlated with the Elo scores representing human judgment for the segmentations of `Park' or `Mushrooms'. However, BSM distances showed a better correlation when applied to the `Bottle' segmentations. These findings suggest that the LAD and MADLAD metrics represent an improvement over BSM and NHD in aligning with human perceptions of segmentation differences.

It is important to recognize potential limitations in the human judgment reflected by the Elo scores, particularly in the perception of differences in the quality of the `Bottle' image segmentations. These segmentations tended to be more challenging to distinguish from one another compared to the other two images, which could affect the reliability of the judgments.

Given the subjective nature of human judgment of image segmentation quality, conducting further evaluations could enhance the robustness of these findings. It would be valuable to repeat the experiment with more participants, potentially including the same participants at different times, and to expand the selection of images. This would help to confirm the consistency of the results and provide a more comprehensive assessment of the segmentation metrics' performance.

\subsection{Application of LAD and MADLAD in Genetic Algorithm}
\label{sec:example_app}

The final experiment will explore the application of LAD and MADLAD metrics in a ``real world" scenario, designed to assist scientific researchers in image understanding workflows such as point selection, region segmentation, and counting. The goal of SEE-Segment, a project at SEE-Insight, is to create a tool that transcends traditional annotation systems \cite{colbry_toward_2023, see-segment}. As scientists annotate their images, our tool will leverage their annotations from the very first image to navigate the ``algorithm space" using a genetic algorithm. This process aims to identify candidate algorithms that align with their specified workflows. If a suitable candidate is found, the tool will suggest this algorithm to the researcher. At a minimum, using such a tool will require no more time than manual annotation, providing researchers with a valuable annotated dataset for further scientific use or integration into traditional ML systems. Optimally, the tool could identify effective algorithms automatically, potentially accelerating parts of the annotation process, thus saving time and enhancing the overall research workflow.

\begin{figure}[h]
	\begin{center}
		\includegraphics[width=0.8\textwidth]{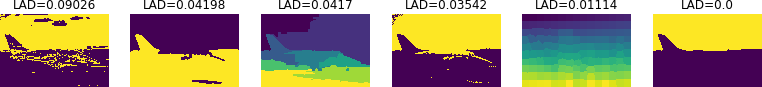}
		\includegraphics[width=0.8\textwidth]{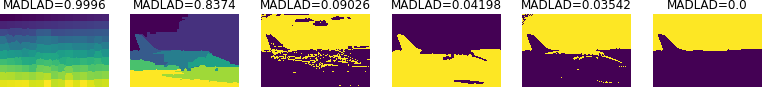}
		\caption{Segmentation quality spectrum: Images are ordered from left to right, from the poorest to the best segmentation quality based on LAD and MADLAD metric values, as demonstrated on an image from the Sky dataset \cite{alexandre_ift-slic_2017}. The ground truth is displayed on the far right for reference. Each image's color variation reflects the number of distinct regions identified by different segmentation algorithms. A robust distance metric should yield consistent evaluations irrespective of the labeling variations used in the segmentations.}
    	\label{fig:SKY}
	\end{center}
\end{figure}

\begin{figure}[h]
	\begin{center}
		\includegraphics[width=0.8\textwidth]{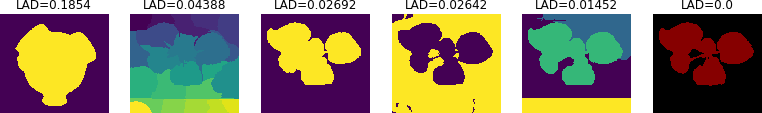}
		\includegraphics[width=0.8\textwidth]{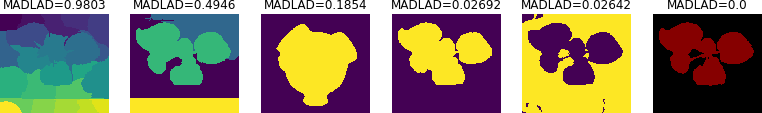}
		\caption{LAD and MADLAD metrics were applied to segmentations of an image from the KOMATSUNA dataset \cite{uchiyama_easy--setup_2017}, demonstrating segmentation with a binary label array. The segmentations are ordered from the least to the most accurate, ending with the ground truth displayed as a reference on the far right.}
    	\label{fig:Plant}
	\end{center}
\end{figure}

\begin{figure}[h]
	\begin{center}
		\includegraphics[width=0.8\textwidth]{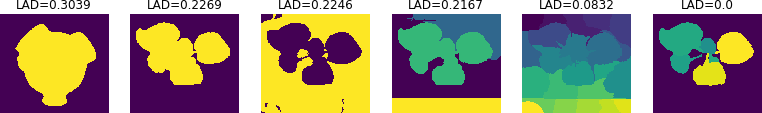}
		\includegraphics[width=0.8\textwidth]{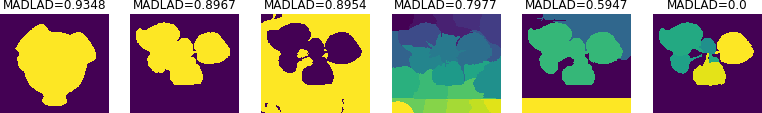}
		\caption{LAD and MADLAD metrics were applied to segmentations of an image from the KOMATSUNA dataset \cite{uchiyama_easy--setup_2017}, here using a multi-label array. The segmentations are arranged from the least to the most accurate, with the ground truth positioned as a reference on the far right.}
    	\label{fig:Multi_Plant}
	\end{center}
\end{figure}

In this context, LAD and MADLAD metrics guide a genetic search algorithm through a high-dimensional, non-differentiable search space of segmentation algorithms and their hyper-parameters. To demonstrate the strengths and limitations of these metrics, a variety of segmentation solutions are presented in Figures \ref{fig:SKY}, \ref{fig:Plant}, and \ref{fig:Multi_Plant}. These solutions were selected to illustrate the approach's capabilities and limitations rather than the quality of individual solutions.

Figure \ref{fig:SKY} displays examples of LAD and MADLAD measurements applied to various segmentations of an airplane against the sky from the Sky dataset \cite{alexandre_ift-slic_2017}. The images are ordered based on their distance metrics from the highest on the left, representing the poorest segmentations, to the lowest on the right. The rightmost images, which are identical to the ground truth, achieve LAD and MADLAD values of zero, indicating a perfect match.

Notably, the image with the second lowest LAD value (second from the right) contains numerous regions, a stark contrast to the ground truth that consists only of two simple regions: object and background. This results in a misleadingly low LAD value, which might suggest a high-quality segmentation despite its impracticality for certain applications. The presence of numerous regions, though reducing the distance metric, might not align with specific application needs, demonstrating the importance of a metric that yields consistent results regardless of label variations.

The MADLAD metric offers a potentially more suitable solution by providing results with a comparable or reduced number of regions. This adjustment better aligns with practical needs where fewer, more clearly defined regions are advantageous, demonstrating MADLAD's relative adaptability to the context of the task.

Figure \ref{fig:Plant} illustrates LAD and MADLAD measurements applied to an image from the KOMATSUNA dataset \cite{uchiyama_easy--setup_2017}, with the binary ground truth positioned at the far right. The images are ordered from highest to lowest metric values, starting with the poorest examples on the far left and culminating in the ground truth. Notably, a very poor segmentation (far left under the LAD metric) shifts towards the right under MADLAD. This shift occurs because, despite the low quality of the segmentation, MADLAD favors solutions with a similar number of regions to the ground truth, showcasing its bias towards matching region counts.

Figure \ref{fig:Multi_Plant} presents results from the KOMATSUNA dataset \cite{uchiyama_easy--setup_2017}, this time using a multi-label ground truth. The images are organized from the poorest to the best segmentation, concluding with the ground truth displayed on the far right for reference. While no solution perfectly matches the ground truth, the visible progression in segmentation quality suggests that many of the results would be adequate for various practical applications.

\section{Discussion and Conclusion}

Creating effective and informative distance metrics is crucial for automating the search through segmentation algorithms. In this study, we introduced LAD and MADLAD, innovative metrics designed to evaluate multi-label image segmentation. These metrics address the challenges of comparing segmented images to their corresponding ground truths, where traditional metrics often fall short. While acknowledging the inherent biases in any metric, LAD and MADLAD enable us to quantify the performance of segmentation algorithms within our search space in a way that is tailored to specific image segmentation problems.

Both metrics have demonstrated correlations with human judgment, particularly in their ability to assess consistency across diverse segmentation scenarios. MADLAD is especially effective in adapting to the number of regions in segmentations, aligning closely with human evaluative standards and practical needs.

Our research underscores that while no metric perfectly matches human judgment, striving for metrics that emulate human-like discernment is crucial, especially in complex scenarios such as those in the multi-label segmentation. MADLAD's attention to region counts proves particularly beneficial in these cases. These metrics also facilitate the automation of searching through segmentation algorithms by quantifying each algorithm's performance relative to specific segmentation challenges. 

Future applications of LAD and MADLAD aim to streamline scientific research workflows by reducing the time spent on manual annotations and enhancing the precision of automated systems. To expand these metrics' utility, further research should explore a broader range of images and segmentation challenges, testing these metrics under varied conditions. Additionally, refining these metrics to minimize biases and better align with practical segmentation needs remains a priority. Engaging more evaluators in the segmentation assessment process will enrich our understanding and help fine-tune these metrics for optimal human perceptual accuracy. These metrics will also be employed to assess the sensitivity of segmentation algorithms, measuring the extent to which output variations correspond to changes in input hyper-parameters.

In summary, LAD and MADLAD represent significant advancements in the computational assessment of image segmentation, providing essential tools for researchers and practitioners aiming to enhance the efficiency and accuracy of image-based analysis. This marks a promising direction for ongoing developments in image processing technology.

\bibliography{Elcvia2e} 

\begin{thebibliography}{10}
\providecommand{\url}[1]{#1}
\csname url@samestyle\endcsname
\providecommand{\newblock}{\relax}
\providecommand{\bibinfo}[2]{#2}
\providecommand{\BIBentrySTDinterwordspacing}{\spaceskip=0pt\relax}
\providecommand{\BIBentryALTinterwordstretchfactor}{4}
\providecommand{\BIBentryALTinterwordspacing}{\spaceskip=\fontdimen2\font plus
\BIBentryALTinterwordstretchfactor\fontdimen3\font minus \fontdimen4\font\relax}
\providecommand{\BIBforeignlanguage}[2]{{%
\expandafter\ifx\csname l@#1\endcsname\relax
\typeout{** WARNING: IEEEtran.bst: No hyphenation pattern has been}%
\typeout{** loaded for the language `#1'. Using the pattern for}%
\typeout{** the default language instead.}%
\else
\language=\csname l@#1\endcsname
\fi
#2}}
\providecommand{\BIBdecl}{\relax}
\BIBdecl

\bibitem{Jaccard}
P.~Jaccard, ``The distribution of the flora in the alpine zone,'' \emph{New Phytologist}, vol.~11, pp. 37--50, 1912.

\bibitem{choi_survey_2010}
S.-s. Choi and S.-h. Cha, ``A survey of {Binary} similarity and distance measures,'' \emph{Journal of Systemics, Cybernetics and Informatics}, pp. 43--48, 2010.

\bibitem{IoU}
E.~Shelhamer, J.~Long, and T.~Darrell, ``Fully convolutional networks for semantic segmentation,'' \emph{IEEE Conference on Computer Vision and Pattern Recognition (CVPR)}, 2015.

\bibitem{Dice}
L.~R. Dice, ``Measures of the amount of ecologic association between species,'' \emph{Ecology}, vol.~26, pp. 297--302, 1945.

\bibitem{VNet}
F.~Milletari, N.~Navab, and S.-A. Ahmadi, ``{V-Net}: Fully convolutional neural networks for volumetric medical image segmentation,'' \emph{Fourth International Conference on 3D Vision (3DV)}, 2016.

\bibitem{hamming_error_1950}
R.~Hamming, ``Error correcting codes,'' \emph{Bell System Technical Journal}, vol.~29, pp. 147--60, 1950.

\bibitem{Hausdorff}
F.~Hausdorff, \emph{Grundzüge der Mengenlehre}.\hskip 1em plus 0.5em minus 0.4em\relax Von Veit \& Company, Leipzig, 1914.

\bibitem{Matthews}
B.~W. Matthews, ``Comparison of the predicted and observed secondary structure of {T4} phage lysozyme,'' \emph{Biochimica et Biophysica Acta (BBA) - Protein Structure}, vol. 405, pp. 442--451, 1975.

\bibitem{PR}
J.~Davis and M.~Goadrich, ``The relationship between precision-recall and {ROC} curves,'' \emph{ICML '06: Proceedings of the 23rd international conference on Machine learning}, pp. 233--240, 2006.

\bibitem{sys}
M.~Sokolova and G.~Lapalme, ``A systematic analysis of performance measures for classification tasks,'' \emph{Information Processing \& Management}, pp. 427--437, 2009.

\bibitem{ROC}
J.~P. Egan, \emph{Signal detection theory and ROC-analysis}.\hskip 1em plus 0.5em minus 0.4em\relax Academic Press, New York, 1975.

\bibitem{AUC}
J.~A. Hanley and B.~J. McNeil, ``The meaning and use of the area under a receiver operating characteristic {(ROC)} curve,'' \emph{Radiology}, vol. 143, pp. 29--36, 1982.

\bibitem{fast}
J.~Lewis, ``Fast normalized cross-correlation,'' \emph{Industrial Light and Magic}, 1995.

\bibitem{fft}
B.~Reddy and B.~Chatterji, ``An {FFT}-based technique for translation, rotation, and scale-invariant image registration,'' \emph{IEEE Transactions on Image Processing}, vol.~5, pp. 1266--1271, 1996.

\bibitem{mustafa_modified_2015}
\BIBentryALTinterwordspacing
A.~A.~Y. Mustafa, ``A modified hamming distance measure for quick detection of dissimilar binary images,'' in \emph{International {Conference} on {Computer} {Vision} and {Image} {Analysis} {Applications}}.\hskip 1em plus 0.5em minus 0.4em\relax Sousse, Tunisia: IEEE, Jan. 2015, pp. 1--6. [Online]. Available: \url{http://ieeexplore.ieee.org/document/7351897/}
\BIBentrySTDinterwordspacing

\bibitem{miou}
L.~C. Chen, G.~Papandreou, I.~Kokkinos, K.~Murphy, and A.~L. Yuille, ``{DeepLab}: Semantic image segmentation with deep convolutional nets, atrous convolution, and fully connected {CRFs},'' \emph{IEEE Transactions on Pattern Analysis and Machine Intelligence}, vol.~40, pp. 834--848, 2018.

\bibitem{FMI}
E.~Fowlkes and C.~Mallows, ``A method for comparing two hierarchical clusterings,'' \emph{Journal of the American Statistical Association}, vol.~78, pp. 553--569, 1983.

\bibitem{Tversky}
A.~Tversky, ``Features of similarity,'' \emph{Psychological Review}, vol.~84, pp. 327--352, 1977.

\bibitem{tvloss}
S.~S.~M. Salehi, D.~Erdogmus, and A.~Gholipour, ``Tversky loss function for image segmentation using {3D} fully convolutional deep networks,'' \emph{International Workshop on Machine Learning in Medical Imaging}, 2017.

\bibitem{alexandre_ift-slic_2017}
\BIBentryALTinterwordspacing
E.~B. Alexandre, ``\BIBforeignlanguage{pt-br}{{IFT}-{SLIC}: geração de superpixels com base em agrupamento iterativo linear simples e transformada imagem-floresta},'' text, Universidade de São Paulo, Jun. 2017. [Online]. Available: \url{http://www.teses.usp.br/teses/disponiveis/45/45134/tde-24092017-235915/}
\BIBentrySTDinterwordspacing

\bibitem{uchiyama_easy--setup_2017}
H.~Uchiyama, S.~Sakurai, M.~Mishima, D.~Arita, T.~Okayasu, A.~Shimada, and R.-i. Taniguchi, ``An easy-to-setup {3D} phenotyping platform for {KOMATSUNA} dataset,'' in \emph{The {IEEE} {International} {Conference} on {Computer} {Vision} ({ICCV}) {Workshops}}, Oct. 2017.

\bibitem{colbry_toward_2023}
\BIBentryALTinterwordspacing
D.~Colbry, ``Toward an automatic exploration of algorithm space to speed up image annotation for applications in scientific image understanding,'' in \emph{2023 {IEEE} {Applied} {Imagery} {Pattern} {Recognition} {Workshop} ({AIPR})}, Sep. 2023, pp. 1--6, iSSN: 2332-5615. [Online]. Available: \url{https://ieeexplore.ieee.org/document/10440694}
\BIBentrySTDinterwordspacing

\bibitem{see-segment}
\BIBentryALTinterwordspacing
------, ``{SEE-Segment},'' 2024. [Online]. Available: \url{https://github.com/see-insight/see-segment}
\BIBentrySTDinterwordspacing

\bibitem{gimp}
``{GIMP} - {GNU} {Image Manipulation Program},'' \url{https://www.gimp.org/}, accessed: 2023-08-08.

\bibitem{github-see-segment}
\BIBentryALTinterwordspacing
D.~Colbry, ``{SEE-Segment Examples},'' 2024. [Online]. Available: \url{https://github.com/see-insight/see-segment/tree/master/Image_data/Examples}
\BIBentrySTDinterwordspacing

\bibitem{elo}
\BIBentryALTinterwordspacing
A.~E. Elo, ``The proposed {USCF} rating system, its development, theory, and applications,'' \emph{Chess Life}, vol. XXII, no.~8, pp. 242--247, 1967. [Online]. Available: \url{https://uscf1-nyc1.aodhosting.com/CL-AND-CR-ALL/CL-ALL/1967/1967_08.pdf#page=26}
\BIBentrySTDinterwordspacing

\bibitem{tkinter}
{Tkinter}, ``tkinter — {Python interface to Tcl/Tk},'' \url{https://docs.python.org/3/library/tkinter.html}, 2024, accessed on 5/6/2024.

\bibitem{pillow}
{Jeffrey A. Clark (Alex) and contributors}, ``Pillow - {Python Imaging Library (Fork)},'' \url{https://pypi.org/project/Pillow/}, 2024, accessed on 5/6/2024.

\end{thebibliography}
\bibliographystyle{IEEEtran}






\end{document}